\documentclass{article}
\usepackage{spconf,amsmath,graphicx}

\usepackage{comment}
\usepackage{booktabs}
\usepackage{placeins}
\usepackage{float}
\usepackage[usenames,dvipsnames]{xcolor}
\usepackage[super]{nth}

\title{CCC-wav2vec 2.0: Clustering aided Cross Contrastive Self-supervised learning of speech representations}
\name{Vasista Sai Lodagala$^1$, Sreyan Ghosh$^2$, S. Umesh$^1$}
\address{$^1$Indian Institute of Technology, Madras\\  $^2$University of Maryland, College Park}
\copyrightnotice{978-1-6654-7189-3/22/\$31.00~\copyright2023 IEEE}
\begin{document}
%
\maketitle
\begin{abstract}
While Self-Supervised Learning has helped reap the benefit of the scale from the available unlabeled data, the learning paradigms are continously being bettered. We present a new pre-training strategy named ccc-wav2vec 2.0, which uses clustering and an augmentation based cross-contrastive loss as its self-supervised objective. Through the clustering module we scale down the influence of those negative examples that are highly similar to the positive. The Cross-Contrastive loss is computed between the encoder output of the original sample and the quantizer output of its augmentation, and vice-versa, bringing robustness to the pre-training strategy. ccc-wav2vec 2.0 achieves upto 15.6\% and 12.7\% relative WER improvement over the baseline wav2vec 2.0 on the test-clean and test-other sets respectively of LibriSpeech, without the use of any language model. The proposed method also achieves upto 14.9\% relative WER improvement over the baseline wav2vec 2.0, when fine-tuned on Switchboard data.
\end{abstract}
\begin{keywords}
self-supervised learning, automatic speech recognition, domain adaptation, telephone speech
\end{keywords}

\vspace{-0.5em}
\section{Introduction}
\vspace{-0.5em}

The use of SSL to learn high-level representations from unlabeled data has received much attention in the last few years in the domains of Computer Vision (CV) \cite{chen2020simple}, Natural Language Processing (NLP), \cite{47751} and Spoken Language Processing (SLP) \cite{baevski2020wav2vec}. Though much progress has been made in CV and NLP, self-supervised learning for SLP has been relatively understudied. In a majority of prior work, SSL for SLP solves a variant of Masked Acoustic Modeling (MAM), either through instance discrimination using contrastive learning \cite{baevski2020wav2vec,chen2022unispeech}, or masked prediction \cite{hsu2021hubert,9814838}. There is, however, much potential to improve the existing self-supervised tasks for better representation learning. In this paper, we introduce two improvements over the standard wav2vec 2.0, which help learn better and more robust speech representations through SSL.

Though SSL has been seen to benefit from scale \cite{hannun2021history}, the performance of SSL for speech with limited unlabeled data needs further attention. Considering that even unlabeled data in most languages is limited in real-world scenarios, SSL algorithms that can learn useful representations even in low-resource regimes are the need of the hour \cite{hannun2021history}. Data augmentation has proven to be an effective strategy for supervised learning setups \cite{ko2015audio,amodei2016deep} when the amount of labeled data is limited. Very recently, self-supervised learning in speech has also shown to benefit from data augmentation \cite{sriram2022wav2vec,ieee_aug}. \cite{kharitonov2021data} showed that data augmentation benefits Contrastive Predictive Coding \cite{oord2018representation} when a limited amount of unlabeled data is available. \cite{ieee_aug} shows how introducing specific augmentations makes their speech recognition model more robust to far-field, multi-talker noisy environments. Contrastive learning to learn representations that maximize the agreement between differently augmented views of the same data is a methodology predominant in CV and has achieved state-of-the-art results in many applications \cite{chen2020simple,caron2020unsupervised}. Inspired by this, we add an auxiliary task to the standard wav2vec 2.0 Contrastive Learning task, wherein we contrast the anchors with negatives generated from an augmented sample and vice-versa. This makes the model more robust to augmentations, which in turn helps learn better representations.

Surprisingly, the choice of negative samples in a Contrastive Learning for SLP setup has drawn much less attention in the literature. 
Very often, given an “anchor” point $x$, “negative samples” $x^{-}_{i}$s are randomly sampled from the training data, independent of how informative they may be for the learned representation. Though very recently CV has seen growing attention to this line of research \cite{robinson2021contrastive,chuang2020debiased,wang2022negative}, to the best of our knowledge, there is no existing work in speech despite many state-of-the-art systems solving a contrastive learning task for self-supervised speech representation learning \cite{baevski2020wav2vec,chen2022unispeech}.

We look at Masked Acoustic Modelling (MAM) from the lens of language modeling and hypothesize that, similar to the Contrastive Learning setups in NLP \cite{bhattacharjee2022text}, it is important to sample negatives that are semantically different. The need amplifies in speech representation learning, where the negative sampling strategy becomes all the more important due to the quasi-stationary nature of speech which makes several consecutive speech frames correspond to the same phone or sound. Moreover, Contrastive Learning models that use instance discrimination as a pre-training task tend to fall into over clustering \cite{wang2021solving} during training.
Thus, we hypothesize that negative examples mapped very close to the anchor in terms of their similarity might represent the same phone or class.
Considering such negative examples would contradict the representation learning task, which should primarily focus on discriminating between sounds or phones.

To arrive at more informative negatives for the contrastive loss, we propose to cluster our potential negative examples and diminish the effect of those negatives in the loss computation that fall into the same cluster as the positive. Simply put, this process identifies the weak non-informative negatives from our population and reduces their impact on the loss computation.

We also demonstrate the robustness of the proposed approach through tasks such as Domain Adaptation and zero-shot decoding on the Switchboard \cite{225858} and Wall Street Journal (WSJ) \cite{paul1992design} datasets, respectively.
To summarize, our primary contributions are as follows:
\begin{itemize}
    \item We introduce an augmentation of the original sample and use its representations to add an auxiliary Cross-Contrastive loss to the existing contrastive loss in wav2vec 2.0.
    \item We demonstrate the usefulness of a clustering module to segregate the negative examples and thereby control the effect of the weak non-informative negative examples in the contrastive learning task.
    \item Combining the above two modules leads to the development of ccc-wav2vec 2.0, a robust pre-training approach that consistently outperforms wav2vec 2.0 in tasks such as ASR, Domain Adaptation, and zero-shot decoding.
\end{itemize}
Our code and models are publicly available on GitHub\footnote{https://github.com/Speech-Lab-IITM/CCC-wav2vec-2.0} \footnote{Correspondence to Vasista Sai Lodagala: vasista.lodagala@gmail.com}.

\vspace{-0.5em}
\section{Related Work}
\vspace{-0.5em}

SSL for speech representation learning has been prevalent in the form of MAM. Most of the MAM approaches introduced in literature aim to either predict the class of the masked entity using a classification objective as in \cite{hsu2021hubert,9814838}, or reconstruct the original frame as in \cite{liu2020mockingjay,liu2021tera}, or enforce similarity between the prediction of the network for the masked frame and a quantized representation of the original masked frame by solving a Contrastive Learning task as in \cite{baevski2020wav2vec}. On the other hand, some of them propose to solve two of these tasks simultaneously \cite{jiang2020speech,chen2022unispeech,chung2021w2v,qian2022contentvec}.

\vspace{-0.5em}
\subsection{Negative Sampling}
\vspace{-0.5em}
Contrastive Learning has been observed to dominate self-supervised speech representation learning methodologies in various forms \cite{baevski2020wav2vec,chen2022unispeech}, constantly achieving new State of the Art (SOTA) results on a variety of SLP tasks \cite{yang2021superb}. However, it is hard to find works that discuss efficient negative mining for Contrastive Learning in SLP. Alternatively, this line of research has seen great success in CV \cite{wang2022negative,robinson2021contrastive,chuang2020debiased,li2021contrastive}. Also, this idea has been successfully applied in metric learning, where most of the works \cite{schroff2015facenet,suh2019stochastic} observe that it is helpful to use negative examples that are difficult to be discriminated from the current embedding. \cite{oord2018representation} was one of the first works to analyze the effect of efficient negative sampling, wherein they observed a drop in performance when negatives were not mined from the same speaker. To the best of our knowledge, this is the first work to design the Contrastive Learning task for speech SSL to explicitly control the choice of negatives.

\vspace{-0.5em}
\subsection{Data Augmentation}
\vspace{-0.5em}
The usefulness of data augmentations for robust SLP has been explored extensively, primarily on a supervised learning setup \cite{ko2015audio,amodei2016deep}. A wide range of SLP tasks like ASR \cite{park2019specaugment,meng2021mixspeech}, Speaker Identification \cite{desplanques2020ecapa} etc., have been seen to benefit from data augmentations. Aspects of low-resource learning \cite{park2019specaugment,meng2021mixspeech}, far-field and noisy environment recognition \cite{ko2017study,tsunoo2021data} have primarily seen to benefit the most with data augmentation. Very recently, the benefits of data augmentation have been explored in SSL-based speech representation learning \cite{sriram2022wav2vec,ieee_aug}, where the former focuses on low-resource and the latter on improving ASR in far-field and noisy environments. In a recent work based on wav2vec 2.0, \cite{ieee_aug} proposes a Multi-Variant Consistency based objective wherein multiple augmented versions of the same audio sample are created. The original audio sample is discarded, and a contrastive loss between the multiple augmented versions is computed. Our proposed approach differs from this work in the following ways: 1) We retain the original audio sample and use the cross-contrastive loss with the augmentation as an auxiliary loss in addition to the original wav2vec 2.0 objective. 2) In our loss computation, the effect of the various negative examples is controlled by the clustering module, leading to an informed contrast with the “anchor” in the contrastive loss.


\vspace{-0.5em}
\section{Methodology}
\vspace{-0.5em}
In the following subsections, we elaborate on the Cross-Contrastive setup and the Clustering module, both of which are key components of the proposed ccc-wav2vec 2.0. Finally, we present how these two components are integrated to form ccc-wav2vec 2.0.

\begin{figure*}[t]
\centering
\includegraphics[width=\textwidth]{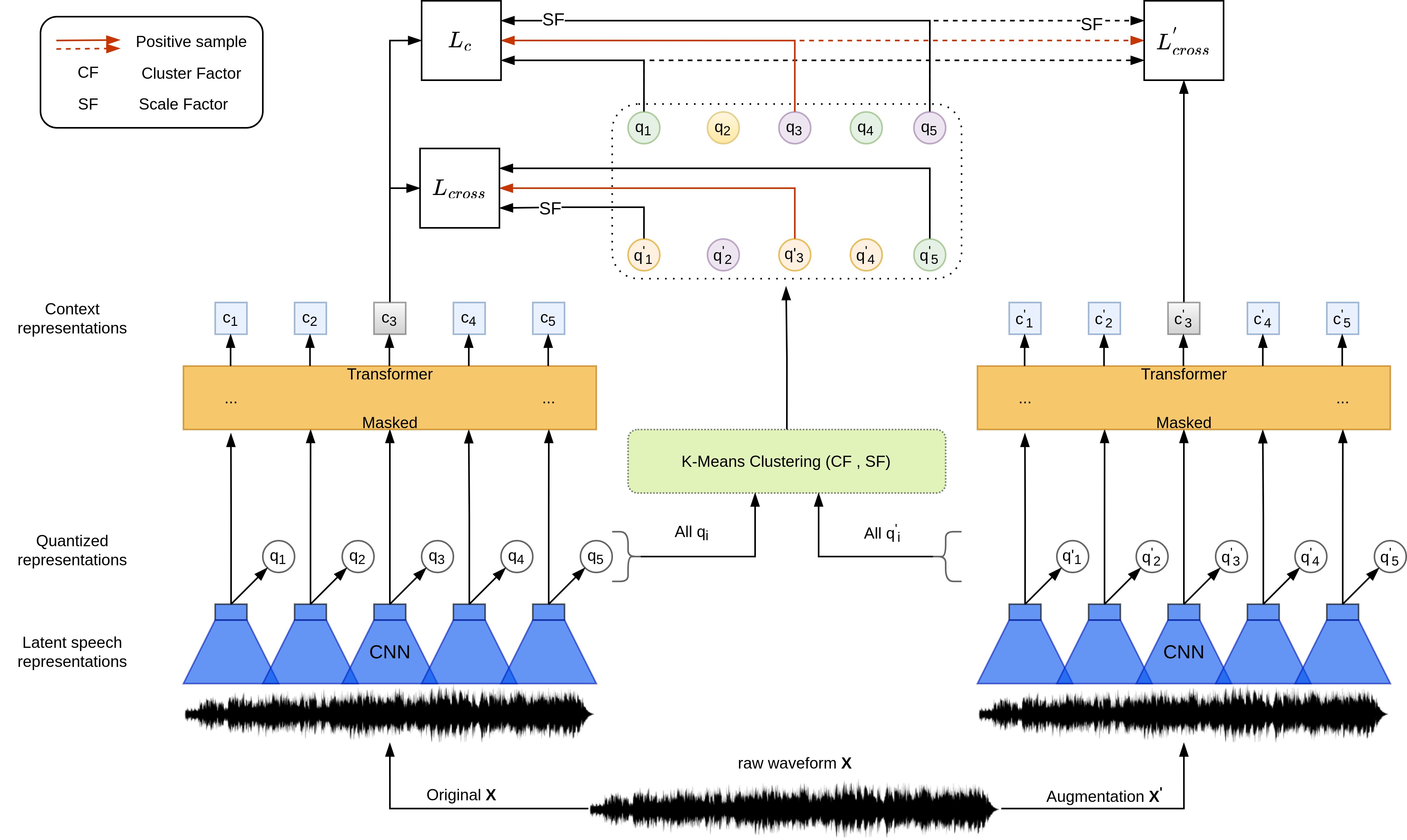}
\caption{Illustration of the ccc-wav2vec 2.0 framework}
\label{fig:CCC}
\end{figure*}

\vspace{-0.5em}
\subsection{Cross-Contrastive Learning}\label{sec:CCL}
\vspace{-0.5em}
Deciphering speech and sound in noisy environments is not a challenging task for humans. However, the same cannot be expected of SSL models, unless they have been trained for it. In order to bring robustness to the pre-training approach, we tap into the augmentations of speech samples.

Given an audio sample $X$, we apply an augmentation over it using the torchaudio-augmentations library \cite{spijkervet_torchaudio_augmentations}, to get $X^{'}$ as the augmented sample. We pass both $X$ and $X^{'}$ through the wav2vec 2.0 model to get the quantized representations $Q_t$, $Q_t^{'}$ and the context representations $C$, $C^{'}$. Unless specified otherwise, the quantized representations are computed only for unmasked latent representations corresponding to the masked regions. The subscript $t$, indicates the masked time steps. An important point to note is that the original and the augmented sample share the same set of masked indices (time-steps) to facilitate the computation of the cross-contrastive loss. In the case of wav2vec 2.0, the standard contrastive loss is defined over each masked time step $t$ as follows:

\begin{equation}
\label{eqn:contr}
    L_c = - log \frac{exp(sim(c_t, q_t)/\kappa)}{\sum_{\tilde{q} \sim Q_t} exp(sim(c_t, \tilde{q})/\kappa)}
\end{equation}
where, $c_t \in C$, $q_t \in Q_t$ and $sim(\mathbf{m},\mathbf{n}) = \mathbf{m}^T\mathbf{n} / \Vert \mathbf{m} \Vert  \Vert \mathbf{n} \Vert$ is used to compute the cosine similarity between context and quantized representations. $\kappa$ represents the temperature parameter. It is to be noted that we do not use all of the frames in $Q_t$ as negative examples. The symbol $\sim$ in the denominator represents the specific number of examples sampled from $Q_t$, to function as distractors or negative examples.

Given the representations from the augmentations, we define the following loss terms:
\begin{equation}
\label{eqn:cross}
    L_{cross} = - log \frac{exp(sim(c_t, q_t^{'})/\kappa)}{\sum_{\tilde{q}^{'} \sim Q_t^{'}} exp(sim(c_t, \tilde{q}^{'})/\kappa)}
\end{equation}
and
\begin{equation}
\label{eqn:cross_1}
    L_{cross^{'}} = - log \frac{exp(sim(c_t^{'}, q_t)/\kappa)}{\sum_{\tilde{q} \sim Q_t} exp(sim(c_t^{'}, \tilde{q})/\kappa)}
\end{equation}
where, $c_t^{'} \in C^{'}$, $q_t^{'} \in Q_t^{'}$. Essentially, the equations \ref{eqn:cross} and \ref{eqn:cross_1} compute the contrastive loss between the context representations of the original sample and the quantized representations of its augmentation, and vice-versa over the masked time steps.

Finally, we define the overall cross-contrastive loss $L_{cc}$ as follows:
\begin{equation}
\label{eqn:cc}
    L_{cc} = \alpha L_c + \beta L_{cross} + \gamma L_{cross^{'}}
\end{equation}
where, $\alpha$, $\beta$ and $\gamma$ are the scalar hyper-parameters that would determine the importance of each of the individual loss terms.

Evidently, the performance of models pre-trained with the $L_{cc}$ loss also depends on the choice of augmentations. The different choices of augmentations are described in Section \ref{sec:results}. It is to be noted that, where ever we mention any variant of contrastive loss, the diversity loss in \cite{baevski2020wav2vec} is also added to the final loss computation.

\vspace{-0.5em}
\subsection{Clustering Module}\label{sec:CM}
\vspace{-0.5em}

As mentioned earlier, the function of the clustering module is to identify the set of weakly informative negatives and diminish their effect in the computation of the contrastive loss.

\vspace{-0.5em}
\subsubsection{Need for Clustering}
\vspace{-0.5em}
As the negative examples are drawn from the quantized representations $Q_t$, we perform clustering over the entire set of frames in $Q_t$. Given that the frames in $Q_t$, are discrete representations computed via product quantization, the need for clustering over $Q_t$ may not be clear immediately. As discussed earlier, the primary motivation behind clustering over quantized representations is that language, spoken or written, is composed of discrete tokens. Negative examples drawn randomly from the population might represent the same class of tokens. In speech, this need is amplified due to its quasi-stationary nature, where several consecutive speech frames could represent the same phone or sound. We mitigate this through negative example elimination and try to get better and more informative negatives by using a ``closeness" measure (in our case, cosine similarity) with the help of clustering.

Another important reason for clustering over quantized representations is that, by default, the product quantization stage of wav2vec 2.0 uses two codebooks, each of size 320. So, the total number of discrete representations possible with this configuration are $320 \times 320 = 102400$. Given such a huge universe of possible discrete representations (coming from product quantization), for any given audio sample $X$, it would be uncommon to find quantized representations $q_t$ in $Q_t$, having the same discrete representation for similar frames. So, to identify such definite similarities, we use a clustering module based on cosine distance. One clear example of such an occurrence is in the case of padded frames. Since the padding is added to samples to create uniformly sized mini-batches during training, the quantized representations of such padded frames can easily be identified and grouped by a clustering module

\vspace{-0.5em}
\subsubsection{Cluster Factor (CF)}\label{sec:CF}
\vspace{-0.5em}
We perform k-means clustering \cite{Jin2010} on $Q_t$ with cosine distance as the metric. The maximum number of iterations is limited to 100. To ensure that the computation speed is not compromised with the introduction of the clustering module, we use the fast-pytorch-kmeans library to perform the clustering on GPUs. The only hyper-parameter to be fixed now is the number of clusters. In our view, this is a hyper-parameter that cannot be fixed easily. This is because speech samples are usually of varied durations. So, padding is introduced in each mini-batch of speech samples to keep the same number of frames per audio. However, the number of frames per audio varies drastically across mini-batches. On the LibriSpeech dataset, we have observed that this number varies between 60 to 380. Given this variance in the number of frames per audio, it is difficult to fix the number of clusters. Doing so could lead to sub-optimal clustering results.

To overcome this hurdle, we introduce a new parameter called cluster factor $(CF)$. Because the clustering is performed separately in each mini-batch, we can fix the number of clusters based on the number of frames per audio in the mini-batch. Given a mini-batch of audio samples, let the number of frames per audio be $NF$, after padding. Then, the number of clusters in the clustering module, for each audio in this mini-batch is given by $ceil(NF/CF)$. It is to be noted that when $CF = 1$, no clustering happens, and there would be no alteration to the standard wav2vec 2.0 architecture.

\vspace{-0.5em}
\subsubsection{Scale Factor (SF)}\label{sec:SF}
\vspace{-0.5em}
Once the clustering has been performed on $Q_t$, we identify the cluster to which the positive sample $q_t$ belongs. Let us denote the cluster to which $q_t$ belongs, by $K$. Our interest then is to control the ``influence" of those negative examples that fall into the same cluster $K$. What is meant by ``influence" here is the effect they have on the computation of contrastive loss as defined in equation \ref{eqn:contr}. As cosine similarity is the metric used over the negative examples, controlling the influence would effectively mean controlling the cosine similarity value. We scale down the cosine similarity of the negative examples in $K$ with $c_t$, by a scaling factor $SF$. Let the sampled set of negatives be represented by the set $Q^{*}$. That is, $Q^{*} = \{ \tilde{q} \sim Q_t \}$. Let the samples in $Q^{*}$ be denoted by $q$. The formula for contrastive loss would then assume the form,
\begin{equation}
\label{eqn:clus_contr}
    L_c = - log \frac{e^{(sim(c_t, q_t)/\kappa)}}{\displaystyle\sum_{q \in K} e^{(sim(c_t, q) . SF/\kappa)} + \displaystyle\sum_{q \not \in K} e^{(sim(c_t, q)/\kappa)}}
\end{equation}

From the construction of equation \ref{eqn:clus_contr}, it is clear that the influence of the negatives sharing the same cluster as the positive example is controlled by the scalar $SF$. When $SF = 1$, equation \ref{eqn:clus_contr} would be the same as the standard contrastive loss as defined in equation \ref{eqn:contr}. However, when $SF = -\infty$, all such negative examples falling in the same cluster as the positive are completely discarded in the loss computation. In our experiments other than the baseline wav2vec 2.0, we vary $SF$ between $-\infty$ and $0.5$. The rationale behind choosing values for $SF$ other than $-\infty$ for certain experiments is as follows: The negative examples in $K$ could sometimes be the hard-negatives \cite{robinson2021contrastive}, as opposed to the common assumption of them being weak non-informative negatives. To account for such possibilities, we experiment with a range of values for $SF$. Models pre-trained using the contrastive loss in equation \ref{eqn:clus_contr} are presented in section \ref{sec:results}, for different choices of $CF$ and $SF$.

\vspace{-0.5em}
\subsection{ccc-wav2vec 2.0}
\vspace{-0.5em}
As illustrated in Fig.\ref{fig:CCC}, the cross-contrastive setup from section \ref{sec:CCL}, and the clustering module from section \ref{sec:CM} are brought together in the design of ccc-wav2vec 2.0. Given the availability of two sets of quantized representations from the original and augmented samples, we have the choice to cluster these representations after pooling all of them.

Once the clustering module clusters the quantized representations from both the samples (original and augmented), it can be seen from Fig.\ref{fig:CCC} that the scaling factor $SF$ is applied to those negative examples belonging to the same cluster as the positive. The clusters can be identified from the shade of color, that the quantized representations bear. For example, $SF$ has been applied on $q_5$ as it belongs to the same cluster as the positive example $q_3$. Similarly, $SF$ has been applied on $q_1^{'}$ as it belongs to the same cluster as the positive example $q_3^{'}$. However, in Fig.\ref{fig:CCC}, $SF$ has not been applied on $q_4^{'}$ though it belongs to the same cluster as $q_3^{'}$. This is because, though $q_4^{'}$ is a potential negative example, it has not been sampled in the negative sampling process. As it does not feature in the loss computation, no scaling factor is applied to it.

The loss computation for ccc-wav2vec 2.0 would be the same as that mentioned in the equations \ref{eqn:contr},\ref{eqn:cross},\ref{eqn:cross_1} and \ref{eqn:cc} of section \ref{sec:CCL}, but with the inclusion of the scaling factor $SF$ in each of those equations, as detailed in section \ref{sec:CM} using equation \ref{eqn:clus_contr}. Models pre-trained using the ccc-wav2vec 2.0 architecture are presented in section \ref{sec:results}, for different configuration choices.

\vspace{-0.5em}
\section{Experimental Setup}
\vspace{-0.5em}

\subsection{Pre-training}
\vspace{-0.5em}
All the models mentioned in tables \ref{tab:aug},\ref{tab:clus} and \ref{tab:ccc} are wav2vec 2.0 BASE models pre-trained on the 360-hour split of the LibriSpeech dataset \cite{panayotov2015librispeech} using the fairseq toolkit\cite{ott2019fairseq}. Owing to the compute resource constraints, the entire 960 hours couldn't be used for pre-training. The models have been pre-trained for 26200 updates, which corresponds to 100 epochs over LibriSpeech-360h, on 4 A-100 GPUs. The number of warmup updates have been set to 19650 and the polynomial decay learning rate scheduler has been used. Both the baseline and the proposed SSL model have the same number of model parameters. Rest of the parameters follow standard configurations made available through \cite{baevski2020wav2vec, ott2019fairseq}.

\vspace{-0.5em}
\subsection{Fine-tuning}
\vspace{-0.5em}
Fine-tuning the pre-trained wav2vec 2.0 models is performed by adding a randomly initialized output layer on top of the Transformer to predict characters \cite{baevski2020wav2vec}. CTC loss has been used during the fine-tuning stage. 

The pre-trained models from Tables \ref{tab:aug}, \ref{tab:clus} and \ref{tab:ccc} are fine-tuned for 36400 updates or 300 epochs on the LibriSpeech-100h split, on 4 A-100 GPUs. 

We also fine-tune these pre-trained models for 11000 updates or 610 epochs, on the 30-hour split of the Switchboard dataset. Fine-tuning the models on Switchboard data which is telephonic speech (out-of-domain), would showcase the robustness of ccc-wav2vec 2.0's pre-training approach.

\begin{table}[!h]
\begin{center}
  \caption{Effect of different augmentations}

  \label{tab:aug}
 
  \begin{tabular}{l l l l l l}
    \toprule
    Model & & \multicolumn{2}{c}{dev} & \multicolumn{2}{c}{test}\\
    \cline{3-4} \cline{5-6}
    & & clean & other & clean & other\\ 
    
    \toprule
    Baseline wav2vec 2.0 & & 12.3 & 30.6 & 12.8 & 31.6\\
     & & & & & \\
    Augmentation I & & 11.1 & 28.5 & 11.5 & 29.6\\
    Augmentation II (*) & & 11.5 & 28.8 & 12.1 & 30.0\\
    Augmentation II & & \bf11.1 & \bf28.1 & \bf11.5 & \bf29.2\\
    
    \bottomrule
    
  \end{tabular}
  
\end{center}
\end{table}

\begin{table}[!h]
\begin{center}
  \caption{Effect of different Clustering choices}

  \label{tab:clus}
 
  \begin{tabular}{l l l l l l}
    \toprule
    Model & & \multicolumn{2}{c}{dev} & \multicolumn{2}{c}{test}\\
    \cline{3-4} \cline{5-6}
    & & clean & other & clean & other\\ 
    
    \toprule
    Baseline wav2vec 2.0 & & 12.3 & 30.6 & 12.8 & 31.6\\
     & & & & & \\
    CF (8), SF ($-\infty$) & & \bf11.3 & 29.7 & 11.9 & 30.3\\
    CF (8), SF (0.1) & & 12.1 & 30.2 & 12.6 & 31.5\\
    CF (8), SF (0.3) & & 11.6 & 29.6 & 12.1 & 30.7\\
    CF (8), SF (0.5) & & 11.3 & 29.6 & 11.9 & 30.5\\
     & & & & & \\
    CF (16), SF ($-\infty$) & & 11.5 & 29.5 & 11.9 & 30.5\\
    CF (16), SF (0.1) & & 12.5 & 30.5 & 12.9 & 31.4\\
    CF (16), SF (0.3) & & 11.4 & \bf29.2 & \bf11.7 & \bf30.3\\
    CF (16), SF (0.5) & & 12.0 & 29.7 & 12.4 & 31.3\\
     & & & & & \\
    CF (24), SF ($-\infty$) & & 11.4 & 29.3 & 12.0 & 30.6\\
    CF (24), SF (0.1) & & 11.8 & 29.7 & 12.3 & 30.7\\
    CF (24), SF (0.3) & & 11.8 & 29.8 & 12.2 & 31.3\\
    CF (24), SF (0.5) & & 11.6 & 30.0 & 12.3 & 31.0\\
    
    \bottomrule
    
  \end{tabular}
  
\end{center}
\end{table}

\begin{table*}
\setlength{\tabcolsep}{7.5pt}
\begin{center}
  \caption{ccc-wav2vec 2.0 performance (WER) over different test sets}

  \label{tab:ccc}

  \begin{tabular}{l l l l l l l l l l l l l}
    \toprule
    Model & & \multicolumn{2}{c}{dev} & & \multicolumn{2}{c}{test} & & \multicolumn{2}{c}{WSJ} & & Switchboard\\
    \cline{3-4} \cline{6-7} \cline{9-10}
    & & clean & other & & clean & other & & dev93 & eval93 & & Dev\\ 
    
    \toprule
     & & & & & & & & & & & & \\
    Baseline wav2vec 2.0 & & 12.3 & 30.6 & & 12.8 & 31.6 & & 31.2 & 31.7 & & 34.9\\
    
    Augmentation II & & 11.1 & 28.1 & & 11.6 & 29.2 & & 29.7 & 29.3 & & 31.8\\
    
    CF (16), SF (0.3) & & 11.4 & 29.2 & & 11.7 & 30.3 & & 29.9 & 29.0 & & 33.2\\
    
    & & & & & & & & & & & & \\
    
    CCC - CF(8), SF(0.3) & & 10.8 & 27.7 & & 11.1 & 28.4 & & 29.7 & 28.5 & & 31.4\\
    
    CCC - CF(8), SF(0.3) - pooled & & 10.5 & 27.1 & & 10.9 & 28.2 & & 29.6 & 28.6 & & 30.6\\
    
    CCC - CF(8), SF(0.5) & & 10.7 & 27.3 & & 11.1 & 28.3 & & 29.4 & 28.3 & & 31.1\\
    
    CCC - CF(8), SF(0.5) - pooled & & 10.6 & 27.3 & & 10.9 & 28.3 & & 29.3 & 28.4 & & 30.5\\
    
    CCC - CF(16), SF(0.3) & & 10.7 & 26.9 & & 11.3 & 28.0 & & 29.1 & 28.8 & & 29.9\\
    
    CCC - CF(16), SF(0.3) - pooled & & \bf10.4 & \bf26.7 & & \bf10.8 & \bf27.6 & & \bf28.9 & \bf28.3 & & \bf29.7\\
    
    CCC - CF(16), SF(0.5) & & 10.5 & 27.1 & & 10.9 & 28.1 & & 29.0 & 28.3 & & 30.4\\
    
    CCC - CF(16), SF(0.5) - pooled & & 10.5 & 27.3 & & 10.9 & 28.0 & & 29.2 & 28.3 & & 30.9\\
    
    \bottomrule
    
  \end{tabular}
  
\end{center}
\end{table*}

\vspace{-0.5em}
\begin{table*}
\setlength{\tabcolsep}{4.25pt}
\footnotesize
\begin{center}
  \caption{Performance of ccc-wav2vec 2.0 pre-trained on LibriSpeech-960h over several downstream tasks from SUPERB \cite{yang2021superb}}

  \label{tab:superb}
  \begin{tabular}{l l l l l l l l l l l l l l l l l}
    \toprule
    Model & Score & KS & IC & PR & ASR & ER & QbE & SF-F1 & SF-CER & SID & SV & SD & SE-STOI & SE-PESQ & SS & ST \\
    \toprule

    ccc-wav2vec2.0 BASE & 1259 & 96.7 & 96.5 & 5.95 & 6.3 & 64.2 & 6.73 & 88.1 & 24.3 & 72.8 & 5.61 & 4.27 & 94.9 & 3.06 & 10.86 & 16.2 \\

    wav2vec2.0 BASE & 674 & 96.2 & 92.4 & 5.74 & 6.43 & 63.4 & 2.33 & 88.3 & 24.8 & 75.2 & 6.02 & 6.08 & 93.9 & 2.55 & 9.77 & 14.8 \\
    
    \bottomrule 
  \end{tabular}
\end{center}
\end{table*}

\vspace{-0.5em}
\section{Results and Analysis} \label{sec:results}
\vspace{-0.5em}

The models from Table \ref{tab:aug} have been pre-trained using the loss in equation \ref{eqn:cc}. For the baseline wav2vec 2.0, the $\alpha$, $\beta$ and $\gamma$ parameters in equation \ref{eqn:cc} have been set to $1$, $0$ and $0$ respectively. Augmentation I in Table \ref{tab:aug} refers to randomly cropping 25\% of the original audio sample and replacing the cropped region with zeroes. Augmentation II, on the other hand, is a combination of 3 different data augmentations. With a probability of $0.6$, additive noise is added to the audio sample at a random signal-to-noise ratio (SNR) between 3 and 15dB. Following this, with a probability of $0.7$, the audio signal is convolved with a random Reverberation Impulse Response (RIR). Finally, with a probability of $0.8$, background noise has been added from random noise samples from the noise set of the MUSAN corpus\cite{musan2015}, at a random signal-to-noise ratio (SNR) between 0 and 15dB. The choice of this set of augmentations for Augmentation II comes from the work of \cite{https://doi.org/10.48550/arxiv.2010.12715}, which uses data augmentations in a supervised setting.
The Augmentation I and Augmentation II models in Table \ref{tab:aug}, have been pre-trained using the loss mentioned in equation \ref{eqn:cc}, with $\alpha$, $\beta$ and $\gamma$ set to $1$, $0.5$ and $0.5$ respectively, with the cross-contrastive loss aiding the contrastive loss. However, the Augmentation II (*) model has $\alpha$, $\beta$ and $\gamma$ set to $0$, $1$ and $1$ respectively. This experiment has been carried out to demonstrate that, the cross-contrastive loss aiding the contrastive loss would lead to better results compared to having only the cross-contrastive loss. 

The models from Table \ref{tab:clus} have been pre-trained using the loss in equation \ref{eqn:clus_contr}. Experiments in Table \ref{tab:clus} have been carried out to demonstrate the impact of the clustering module. Therefore, no augmentations are involved for the experiments in Table \ref{tab:clus}. The Cluster Factor $(CF)$ as defined in section \ref{sec:CF}, and the Scale Factor $(SF)$ as defined in section \ref{sec:SF}, have been varied to arrive at the best possible clustering configuration. The different choices for $CF$ and $SF$ can be found in Table \ref{tab:clus}. We find that having a cluster factor of 16, and a scale factor of 0.3, results in the best average performance among the different choices made for $CF$ and $SF$. The fact that $SF = 0.3$ out-performs $SF = -\infty$, for $CF = 16$, goes on to prove the presence of hard-negatives, among the weakly non-informative negatives.

Finally, the models from Table \ref{tab:ccc}, prefixed with CCC, have been pre-trained using the loss in the equation \ref{eqn:cc}. However, with the introduction of the clustering module, each loss term in equation \ref{eqn:cc}, has the scaling factor $SF$ included. 
For models in Table \ref{tab:ccc}, the suffix ``pooled" indicates that the quantized representations from the original sample and its augmentation are clustered together. Given the performance of different augmentations in Table \ref{tab:aug}, we choose Augmentation II, for all our experiments involving pre-training of ccc-wav2vec 2.0.

The best configuration of ccc-wav2vec 2.0 from Table \ref{tab:ccc}, when fine-tuned on the LibriSpeech-100h split, achieves up to 15.4\%, 12.7\% and 15.6\%, 12.7\% relative WER improvement compared to wav2vec 2.0, over the dev-clean, dev-other and test-clean, test-other sets of the LibriSpeech dataset respectively. The same model achieves up to 7.3\% and 10.7\% relative WER improvement compared to wav2vec 2.0 over the dev93 and eval93 sets of the WSJ dataset. This shows the efficiency of our pre-training approach over unseen data. Also, the robustness of the pre-training approach is evident from its domain adaptation to the Switchboard data. When fine-tuned on the 30-hour subset of the Switchboard data, the model achieves up to 14.9\% relative WER improvement compared to wav2vec 2.0.

\noindent \textbf{SUPERB evaluation}: The BASE variant of ccc-wav2vec 2.0 has been pre-trained on LibriSpeech-960h and was evaluated over the several downstream speech tasks presented over SUPERB \cite{yang2021superb}. The model is ranked \nth{6} over the Challenge public set of the benchmark, outperforming the wav2vec 2.0 BASE model which has been ranked \nth{12} in the same leaderboard.

\vspace{-1em}
\section{Conclusion and Future Work}
\vspace{-0.5em}
In this paper, we introduce ccc-wav2vec 2.0, a novel SSL-based pre-training approach based on wav2vec 2.0, that improves on the original wav2vec 2.0 discrimination task. Our approach consistently outperforms the wav2vec 2.0 architecture while training the same number of parameters. As a part of future work, we would like to explore techniques to sample negatives that make the instance discrimination task harder to solve and also explore new data augmentation techniques.

\vfill\pagebreak

\bibliographystyle{IEEEbib}
\bibliography{IEEE}

\end{document}